\documentclass[11pt]{article}
\usepackage{booktabs}

\usepackage[final]{latex/acl}

\usepackage{times}
\usepackage{latexsym}

\usepackage[T1]{fontenc}
\usepackage[utf8]{inputenc}

\usepackage{microtype}

\usepackage{inconsolata}

\usepackage{graphicx}
\usepackage{makecell}
\usepackage{xcolor}
\usepackage{amsfonts}
\usepackage{amsmath} 
\usepackage{multirow}
\usepackage{arydshln} % 用于绘制表格中的虚线
\usepackage{colortbl} % 用于表格行高亮背景色
\definecolor{graybg}{RGB}{242,242,242}
\definecolor{oursbg}{RGB}{235,245,255}
\usepackage{subcaption}  % 用于支持子图 (subfigure)
\usepackage{float}
\title{LIRQuant: An Efficient Locally Isolated Rotation Quantization Method for Large Language Models}
\author{Desheng Xu \and Hanglin Li \and Tianchen Song \and Zhiyong Zhao \and Kun Zhan \\
  Li Auto Inc. \\
  \texttt{\{xudesheng,lihanglin,songtianchen,zhaozhiyong,zhankun\}@lixiang.com}}

\begin{document}
\maketitle
\begin{abstract}

Post-Training Quantization (PTQ) enables efficient deployment of Large Language Models (LLMs), yet activation outliers remain a key challenge for low-bit quantization. Existing global transformation methods overlook heterogeneous channel sensitivity to quantization noise, leading to suboptimal accuracy-efficiency trade-offs.
To address this, we propose \textbf{Locally Isolated Rotation Quantization (LIRQuant)}, a sensitivity-aware framework that selectively applies transformations. LIRQuant identifies sensitive channels using a noise-weighted sensitivity metric, applying expressive learnable rotations only to these channels and lightweight fixed rotations to the remaining ones. This decoupled design reduces optimization complexity without sacrificing outlier smoothing capability.
Extensive experiments across the LLaMA-2/3, Mistral, Qwen3, and Mixtral families demonstrate that LIRQuant achieves state-of-the-art accuracy under the challenging W4A4KV4 setting. Meanwhile, LIRQuant calibrates LLaMA-3-8B in only 25 
minutes with 20.3 GB peak GPU memory, achieving the fastest calibration and the lowest memory footprint among recent learnable-rotation methods, and its fused inference kernels deliver up to $2.3\times$ prefill speedup and $3.5\times$ memory saving over FP16.
% Our code is available at xxx.

\end{abstract}

\section{Introduction}
Large Language Models (LLMs) have achieved remarkable capabilities across a wide range of tasks \citep{brown2020language,hendrycks2020measuring}. Models such as LLaMA \citep{touvron2023llama}, GPT \citep{achiam2023gpt}, and Qwen \citep{yang2025qwen3} continue to drive progress toward universal language intelligence. However, their massive parameter counts incur substantial memory and computational costs, posing significant challenges for deployment on resource-constrained devices \citep{samsi2023words,fernandez2025energy,luccioni2024power}. Post-Training Quantization (PTQ) addresses this issue by converting weights and activations into low-precision formats, thereby reducing memory footprint and inference cost with minimal performance degradation \citep{frantar2022gptq,shao2023omniquant,ashkboos2024quik}. Consequently, PTQ has emerged as a critical technique for practical LLM deployment.

Activation outliers remain a major obstacle to low-bit quantization of LLMs \citep{dettmers2022gpt3int8,xiao2023smoothquant,lin2024awq}. Recent studies mitigate activation outliers through various linear transformations, including orthogonal rotations and Hadamard transforms, which smooth activation distributions and improve quantization robustness \citep{ashkboos2024quarot,liu2025spinquant}. Existing rotation-based approaches implicitly adopt a homogeneous transformation strategy, where a single dense transformation is jointly optimized across all channels. Such full-dimensional rotations provide strong distribution reshaping capability, but also introduce substantial optimization complexity and parameter overhead.

Through extensive analysis, we observe that Transformer channels exhibit heterogeneous sensitivity to quantization noise. Specifically, only a small subset of channels is fragile under low-bit quantization, while most channels remain comparatively robust. This suggests that different channels require fundamentally different transformation capacities during quantization. However, existing methods such as SpinQuant \citep{liu2025spinquant} and OSTQuant \citep{hu2025ostquant} optimize dense learnable rotations over all channels indiscriminately. Jointly optimizing full-channel rotations forces sensitive and robust channels to share the same transformation subspace, causing optimization capacity to be unnecessarily allocated to robust channels while introducing cross-channel interference that destabilizes sensitive channels. Consequently, globally shared rotations result in suboptimal accuracy-efficiency trade-offs.

Motivated by these observations, we propose Locally Isolated Rotation Quantization (LIRQuant), a sensitivity-aware rotation framework for LLM quantization. LIRQuant decouples sensitive and robust channels into isolated transformation subspaces according to their quantization sensitivity. Subsequently, it selectively applies highly expressive learnable rotations to high-sensitivity channels, while employing lightweight fixed Hadamard transforms for low-sensitivity ones, thereby substantially reducing optimization costs while preserving quantization accuracy. Extensive experiments across the LLaMA-2/3 families, Mistral-7B, Qwen3-8B, and the MoE model Mixtral-8$\times$7B fully demonstrate the effectiveness of LIRQuant. Under the challenging W4A4KV4 setting, LIRQuant matches or outperforms existing rotation-based methods, such as SpinQuant \citep{liu2025spinquant} and FlatQuant \citep{sun2024flatquant}, while requiring only 25\% of the trainable parameters of a dense rotation. Our main contributions are summarized as follows:

\begin{itemize}
    \item We reveal that quantization sensitivity in Transformer activations is highly heterogeneous across channels, and different channels require different transformation capacities during quantization.
    \item We propose LIRQuant, a sensitivity-aware locally isolated rotation framework. It selectively applies learnable rotations to sensitive channels and lightweight Hadamard transforms to insensitive ones, striking a superior accuracy-efficiency trade-off, and deeply integrates this design with the Transformer architecture.
    \item Extensive experiments across multiple LLMs and quantization settings demonstrate that LIRQuant achieves state-of-the-art quantization performance with the lowest calibration overhead among recent rotation-based methods, and further delivers practical inference speedups via fused block-diagonal kernels.
\end{itemize}

% \begin{itemize}
% \item We reveal that quantization sensitivity in Transformer activations is highly heterogeneous across channels, and show that different channels require substantially different transformation capacities during quantization.

% \item We propose \textbf{LIRQuant}, a sensitivity-aware locally isolated rotation framework that selectively applies learnable rotations to sensitive channels and lightweight Hadamard transforms to insensitive ones, achieving a superior accuracy-efficiency trade-off.

% \item We develop a deployment-friendly offline fusion scheme deeply integrated with Transformer architectures, where tailored transformation strategies are designed for different modules to enable efficient inference without additional overhead.

% \item Extensive experiments across multiple LLMs and quantization settings demonstrate that LIRQuant achieves state-of-the-art quantization performance with substantially lower optimization cost than existing rotation-based methods.
% \end{itemize}

\section{Motivation}
\subsection{Basics of Uniform Quantization}
The deployment of Large Language Models (LLMs) faces severe memory and computational challenges. Post-Training Quantization (PTQ), as one of the most effective compression techniques, can significantly reduce memory footprint and accelerate inference speed by converting floating-point weights and activations into low-precision formats. For an arbitrary floating-point tensor $\boldsymbol{X}$, uniform quantization partitions its continuous numerical range $[x_{\min}, x_{\max}]$ into $2^k-1$ intervals of equal length. The quantization process can be formulated as:
\begin{equation}
Q(\boldsymbol{X}) = \mathrm{round}\left(\frac{\boldsymbol{X}}{s}\right) + z,
\end{equation}
where $s = \frac{x_{\max} - x_{\min}}{2^k - 1}$ is the quantization step size (scale), and $z = -\mathrm{round}\left(\frac{x_{\min}}{s}\right)$ is the zero point. The dequantization process is given by:
\begin{equation}
\hat{\boldsymbol{X}} = s \cdot (Q(\boldsymbol{X}) - z).
\end{equation}
The Mean Squared Error (MSE) introduced by quantization is defined as: 
\begin{equation}
\mathrm{MSE}(\boldsymbol{X}, \hat{\boldsymbol{X}}) = \mathbb{E}\left[(\boldsymbol{X} - \hat{\boldsymbol{X}})^2\right].
\end{equation}
Under the assumption of uniform quantization, the expectation of the quantization error satisfies: 
\begin{equation}
\mathbb{E}[\mathrm{MSE}] \approx \frac{s^2}{12}.
\end{equation}
Consequently, the quantization error is proportional to the square of the quantization step size. As indicated by the above equations, the quantization step size $s$ is the core factor determining the quantization error. When outliers are present in the data, $x_{\max}$ or $x_{\min}$ is significantly stretched by extreme values, causing the quantization step size $s$ to increase substantially. Consequently, since the error is proportional to the square of the step size, this increase in $s$ leads to a quadratic amplification of the overall quantization error.

\subsection{Related Work}

Post-Training Quantization (PTQ) for LLMs. PTQ is an efficient technique for compressing LLMs, primarily categorized into weight-only and weight-activation quantization. To alleviate memory bandwidth bottlenecks, methods such as GPTQ \citep{frantar2022gptq}, AWQ \citep{lin2024awq}, and QuIP \citep{chee2023quip} have made significant progress in weight-only quantization through techniques like error compensation, protecting salient weights, or introducing Hadamard matrices.

\begin{figure}[t]
  \includegraphics[width=\linewidth]{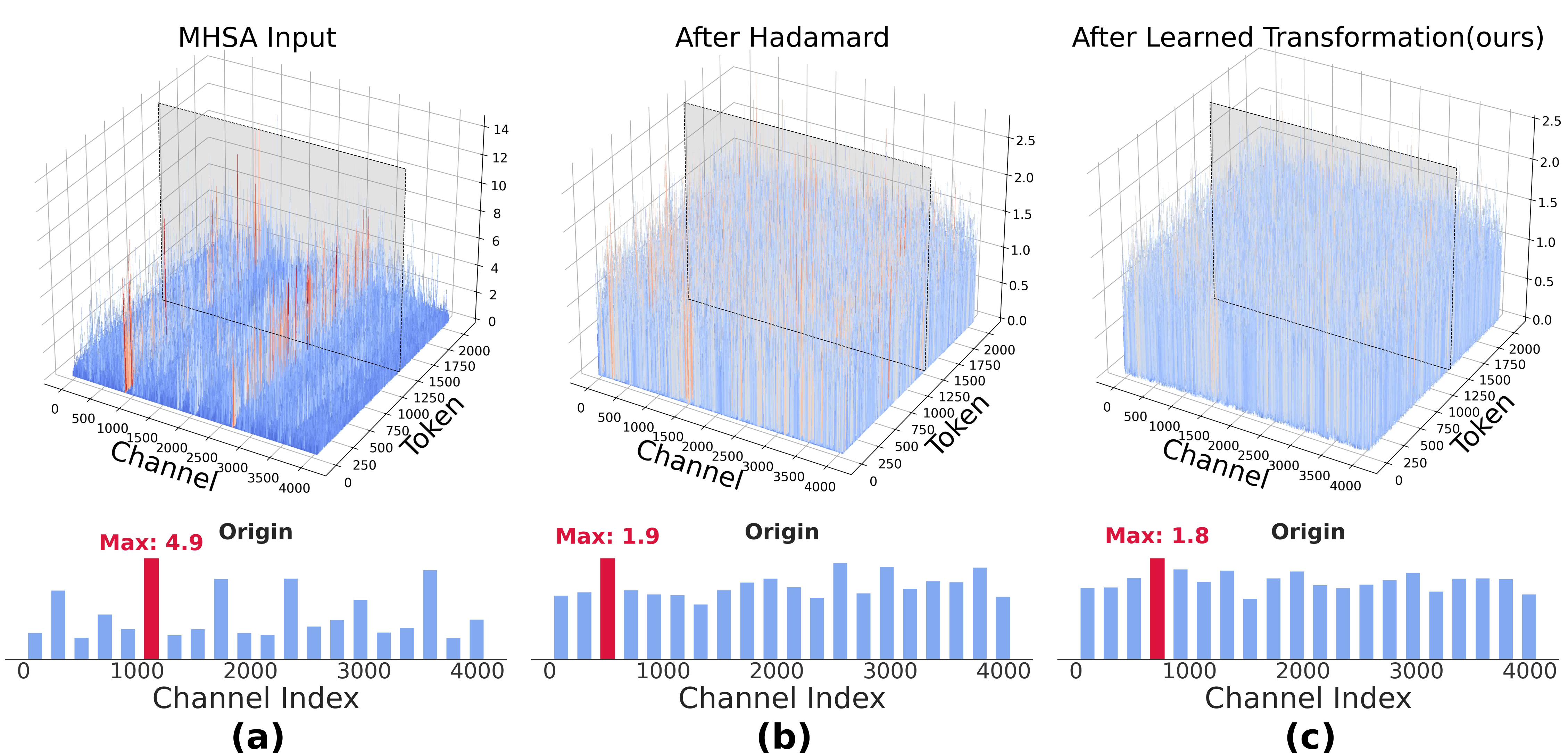}
  \caption{Comparison of outlier suppression effects among different transformation methods. (a) Original activation distribution; (b) Distribution after Hadamard rotation; (c) Distribution after learnable rotation.}
  \label{fig:experiments}
\end{figure}

Unlike weight-only quantization, weight-activation quantization aims to comprehensively accelerate inference. However, the ubiquitous presence of activation outliers in LLMs dominates the dynamic range and is the primary cause of quantization accuracy degradation. To address this issue, early channel-wise scaling methods, such as SmoothQuant \citep{xiao2023smoothquant}, migrate the quantization difficulty from activations to weights, partially mitigating regular outliers. Subsequently, rotation-based methods such as QuaRot \citep{ashkboos2024quarot} and SpinQuant \citep{liu2025spinquant} leverage orthogonal transformations to redistribute activation energy across dimensions, effectively reducing peak magnitudes and improving quantization robustness. Recently, more complex transformation strategies have been proposed: DuQuant \citep{lin2024duquant} introduces a dual transformation combining block-wise rotation and zigzag permutation to tackle massive outliers; OSTQuant \citep{hu2025ostquant} combines orthogonal and scaling transformations to globally optimize the quantization space; and FlatQuant \citep{sun2024flatquant} combines Kronecker decomposition and affine transformations to smooth the distribution while minimizing runtime overhead. As illustrated in Figure~\ref{fig:experiments}, we compare the activation distributions of the MHSA input in the 15th layer of LLaMA-3-8B under different transformation strategies. The original activations contain numerous extreme outliers. Although the Hadamard transform can partially smooth activations, its fixed weights may not be optimal for all activation distributions. In contrast, learning-based methods update the weights of the rotation matrices during training, thus achieving superior suppression of outliers.

% The ubiquitous presence of activation outliers in LLMs is the primary cause of quantization accuracy degradation. Prior works have explored various pre-quantization transformation methods. Early channel-wise scaling approaches, such as SmoothQuant, migrate quantization difficulty from activations to weights and partially mitigate regular outliers. However, they struggle to handle massive activation spikes and often introduce sharper weight distributions. More recent rotation-based methods, including QuaRot and SpinQuant, leverage orthogonal transformations to redistribute activation energy across dimensions. Since orthogonal rotations preserve the overall signal energy while dispersing extreme values across channels, they effectively reduce peak magnitudes and improve quantization robustness.

% As illustrated in Figure~\ref{fig:experiments}, we compare the activation distributions of the MHSA input in the 15th layer of LLaMA-3-8B under different transformation strategies. The original activations contain numerous extreme outliers. Although the Hadamard transform partially smooths activations, its fixed weights may not be the optimal solution for all activation distributions. In contrast, learning-based methods update the weights of the rotation matrices during training, thus achieving superior suppression of outliers.

\subsection{Heterogeneity of Quantization Sensitivity}
Although rotation-based methods substantially improve quantization robustness, existing approaches allocate equal optimization effort and transformation capacity to all channels, uniformly applying dense full-dimensional rotations across the entire activation space. However, this approach fundamentally contradicts the intrinsic heterogeneity of quantization sensitivity in LLM activations, wasting computational resources on low-sensitivity channels that have a negligible impact on model performance.
\begin{figure}[t]
  \centering
  \includegraphics[width=\linewidth]{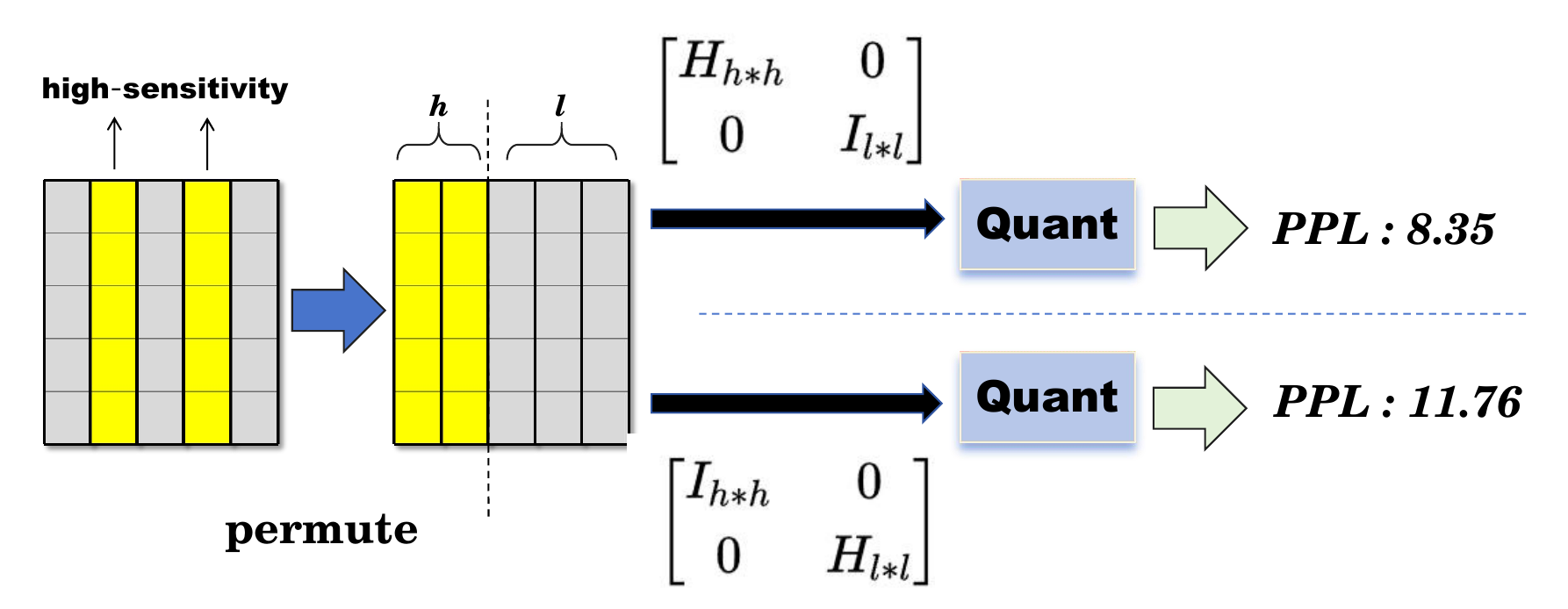}
  \caption{Impact of selectively optimizing sensitive channels under W4A4KV4 quantization.}
  \label{fig:2}
\end{figure}

To validate this conclusion, we conduct a simple yet effective ablation study. As shown in Figure~\ref{fig:2}, on the LLaMA-3-8B model, applying a Hadamard transform to the subspace formed by the top 10\% most sensitive channels of the MHSA input activations, while applying an identity mapping to the remaining channels prior to W4A4KV4 quantization, yields a perplexity (PPL) of 8.35. Conversely, applying an identity mapping to the top 10\% sensitive channel subspace and a Hadamard transform to the remaining channels results in a PPL of 11.76, which is close to the RTN baseline of 11.85. These observations motivate a new insight: rather than transforming all channels indiscriminately, quantization transformations should allocate different transformation capacities based on channel sensitivity. This insight forms the foundation of our proposed LIRQuant.

\section{Method}

\begin{figure*}[t]
  \includegraphics[width=0.99\linewidth]{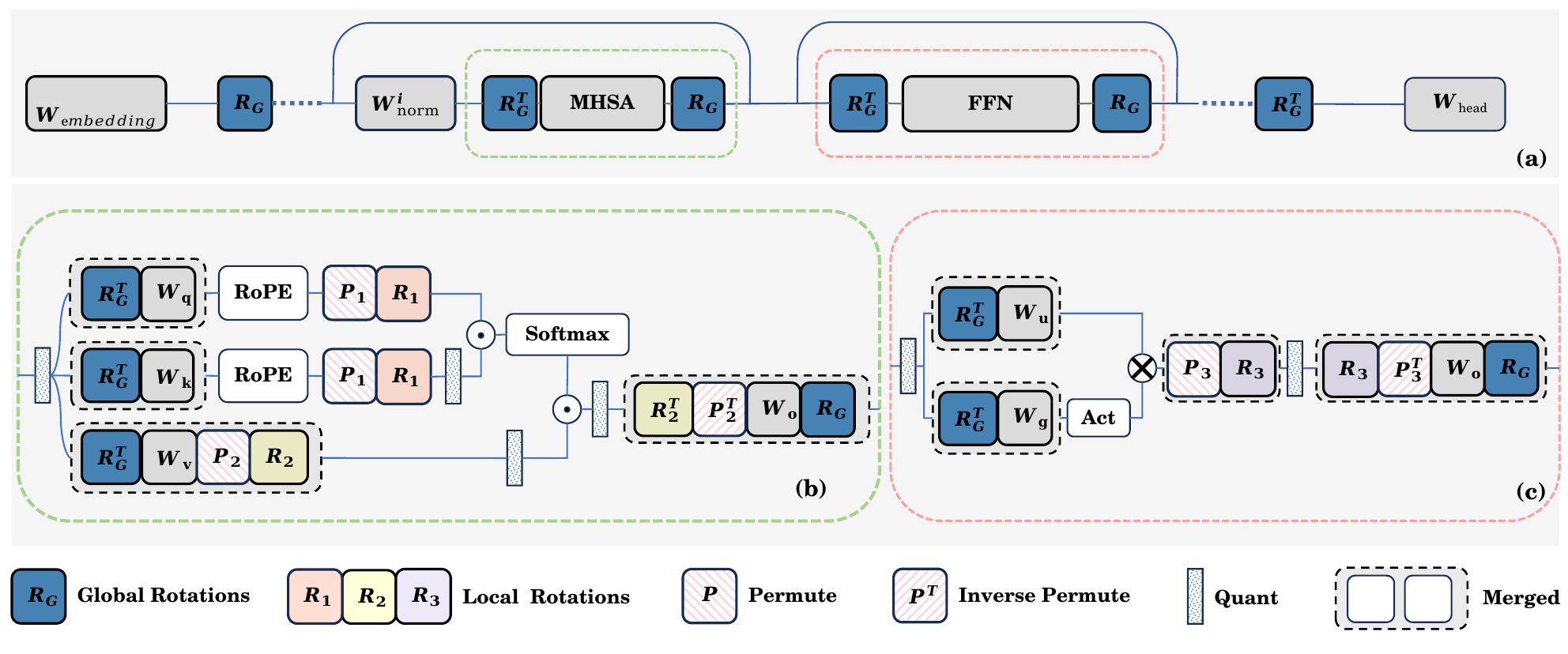}
  \caption{Architectural design of LIRQuant. \textbf{(a)} Global rotation flow across the Transformer blocks. \textbf{(b)} Integration of locally isolated rotations in the MHSA module. \textbf{(c)} Integration of local rotation in the FFN module. Dashed boxes indicate matrices that are fused together to reduce inference overhead.}
  \label{fig:framework}
\end{figure*}

In this paper, we propose LIRQuant, the overall framework of which is illustrated in Figure~\ref{fig:framework}.  By accurately identifying critical subspaces through sensitivity analysis, this method reduces the number of learnable parameters and significantly improves training efficiency, while retaining the core advantages of rotation-based methods in smoothing outliers and facilitating low-bit quantization.

\subsection{Locally Isolated Transformation}

In LIRQuant, we apply an invertible transformation $T$ to activations and weights to smooth outliers. Unlike traditional global dense rotation, LIRQuant introduces a permutation matrix $P \in \{0,1\}^{d\times d}$ sorted by channel sensitivity in descending order, parameterizing the transformation matrix into a block-diagonal form:
\begin{equation}
T = P^{\top}\begin{bmatrix} M_{hi} & 0\\ 0 & H_{lo} \end{bmatrix}P,
\end{equation}
where $M_{hi}$ is a learnable matrix applied to the high-sensitivity subspace $C_{hi}$, and $H_{lo}$ is a fixed Hadamard matrix applied to the low-sensitivity subspace $C_{lo}$. This design aims to maximize the efficiency of the parameters. By allocating learnable parameters exclusively to the critical high-sensitivity channels and applying training-free fixed transformations to the low-sensitivity ones, LIRQuant significantly reduces training overhead while effectively preserving the accuracy of the quantized model.

\subsection{Noise-Weighted Local Functional Sensitivity}
\label{sensitivity}
To accurately partition $C_{hi}$ and $C_{lo}$, we evaluate the sensitivity of each channel based on local functional distortion. For an insertion point $p$, let the intermediate activation be $z_p$ and the local target be $y_p=f_p(z_p)$. The distortion introduced by the quantization perturbation $\hat z_p=z_p+\epsilon_p$ is defined as the expected distance over the calibration set $\mathcal C$:
\begin{equation}
\mathcal D_p=\mathbb E_{\mathcal C}\bigl[D_p\bigl(f_p(z_p+\epsilon_p),f_p(z_p)\bigr)\bigr].
\end{equation}
Through Taylor expansion and Hessian approximation (detailed derivation in Appendix~\ref{sec:sensitivity_derivation}), this distortion can be diagonalized and decomposed as $\mathcal D_p\approx\sum_{j=1}^{d} G_{p,jj}\sigma_{p,j}^2$. Thus, we define the sensitivity of the $j$-th channel as:
\begin{equation}
s_{p,j}=G_{p,jj}\sigma_{p,j}^2.
\end{equation}
This metric decouples the functional sensitivity $G_{p,jj}$ and the quantization noise intensity $\sigma_{p,j}^2$. If weight quantization is also considered, an additive approximation $s_{p,j}=s_{p,j}^{act}+s_{p,j}^{w}$ is adopted.

\subsection{Sensitivity Instantiation}
\label{instantiation}
Based on the above framework, we instantiate the sensitivity $s_{p,j}$ for each insertion point as follows (where $\Delta$ denotes the quantization error):

\begin{itemize}
    \item \textbf{Attention and FFN Inputs:} The local target is the output of the linear layers. The sensitivity of the $j$-th input channel is:
    \begin{equation}
    s^{in}_j=\sum_{r}\bigl(\|\Delta x_j\|_2^2\|w_{r,j}\|_2^2+\|x_j\|_2^2\|\Delta w_{r,j}\|_2^2\bigr)
    \end{equation}
    
    \item \textbf{QK Path:} The target is the attention matrix. Let $A=QK^\top/\sqrt d$ and $P=\operatorname{softmax}(A)$. Using logits-MSE to approximate the local KL distortion, the sensitivity of the $j$-th head channel is:
    \begin{equation}
    s^{qk}_j\approx\frac1d\mathbb E\bigl[\|\Delta q_j\|_2^2\|k_j\|_2^2+\|q_j\|_2^2\|\Delta k_j\|_2^2\bigr]
    \end{equation}
    
    \item \textbf{VO Path:} The target is the value propagation under the fixed full-precision attention weights $P_{fp}$. The sensitivity of the $j$-th value channel is:
    \begin{equation}
    \begin{split}
    s^{vo}_j = \mathbb E\bigl[ & \|P_{fp}\Delta v_j\|_2^2\|w_{o,j}\|_2^2 \\
    & + \|P_{fp}v_j\|_2^2\|\Delta w_{o,j}\|_2^2 \bigr]
    \end{split}
    \end{equation}
    
    \item \textbf{Down Projection Input:} Let $z=\phi(xW_{gate})\odot xW_{up}$, and the target is the final output of the FFN, $zW_{down}$. The sensitivity of the $j$-th channel is:
    \begin{equation}
    \begin{split}
    s^{down}_j &= \|\Delta z_j\|_2^2\|w_{down,j}\|_2^2 \\
    &\quad + \|z_j\|_2^2\|\Delta w_{down,j}\|_2^2
    \end{split}
    \end{equation}
\end{itemize}

\subsection{Overall Workflow and End-to-End Training}

The overall pipeline of LIRQuant consists of two stages:
\begin{enumerate}
    \item \textbf{Sensitivity Analysis and Initialization:} First, we collect activation statistics on a small calibration set and calculate the channel sensitivity $s_{p,j}$ for each layer using the formulas in Section~\ref{instantiation}. Subsequently, based on the determined proportion of high-sensitivity channels, we determine the permutation matrix $P$ and the high-sensitivity subspace dimension $k_p$ for each insertion point.

    \item \textbf{End-to-End Training:} After determining the subspace partition, we freeze the original weights of the LLM and only set the transformation matrices $M_{hi}$ of the high-sensitivity subspaces at each insertion point as learnable parameters. We perform end-to-end training on the calibration set using the standard Cross-Entropy (CE) loss:
    \begin{equation}
    \min_{\{M_{hi}\}} \mathbb{E}_{(X, Y) \in \mathcal{C}} \bigl[ \mathcal{L}_{\text{CE}}\bigl(\text{LLM}_Q(X; \{M_{hi}\}), Y\bigr) \bigr]
    \end{equation}
    where $\mathcal{C}$ is the calibration set, $(X, Y)$ represents the input sequence and its corresponding target labels, $\text{LLM}_Q$ denotes the quantized large language model and locally isolated transformations, and $\{M_{hi}\}$ represents the set of transformation matrices to be optimized across all insertion points. 
\end{enumerate}

\subsection{Integration with the Transformer Architecture}

We now illustrate the integration of our proposed quantization framework with a Transformer block based on the LLaMA architecture. Following conventional practices, we apply low-bit quantization to the inputs and weights of linear projections, while keeping operations such as RMSNorm, RoPE embeddings, and softmax attention scores in FP16.

\textbf{Global Residual Stream.} Our architecture maintains the entire residual stream in a globally rotated space defined by $\mathbf{R}_G$. Specifically, we do not apply permutation to $\mathbf{R}_G$ and instead utilize a full-size learnable matrix for it. As illustrated in Figure~\ref{fig:framework}(a), the initial embedding is multiplied by $\mathbf{R}_G$, and all subsequent Transformer blocks operate within this transformed space. To ensure strict computational equivalence, the learnable scaling parameters of the RMSNorm ($\mathbf{W}_{\text{norm}}^i$) are pre-absorbed into the subsequent projection weights offline. This allows the orthogonal matrix $\mathbf{R}_G$ to commute with the normalization operation without introducing any mathematical discrepancy.

\textbf{Self-Attention.} The self-attention module incorporates the global rotation $\mathbf{R}_G$ and local transformations $\{\mathbf{P}_1, \mathbf{R}_1, \mathbf{P}_2, \mathbf{R}_2\}$. Specifically, the input from the globally rotated residual stream is first projected back using the merged weights $\mathbf{R}_G^\top \mathbf{W}_q$, $\mathbf{R}_G^\top \mathbf{W}_k$, and $\mathbf{R}_G^\top \mathbf{W}_v$. To preserve the positional information, the Rotary Position Embedding (RoPE) is applied \textit{before} any channel-wise transformations. The permutation $\mathbf{P}_1$ and local rotation $\mathbf{R}_1$ are applied to the queries and keys to smooth the attention computation. Since the permutation matrix is inherently an orthogonal matrix, computational equivalence is strictly satisfied. Furthermore, we maintain the computation of the attention map in high precision and quantize the output of the K projection before storing it in the K cache. Similarly, $\mathbf{P}_2$ and $\mathbf{R}_2$ are applied to the values. For the output projection, the inverse transformations $\mathbf{R}_2^\top$ and $\mathbf{P}_2^\top$, along with the global rotation $\mathbf{R}_G$, are analytically absorbed into the output weight $\mathbf{W}_o$, forming the merged weight $\mathbf{R}_2^\top \mathbf{P}_2^\top \mathbf{W}_o \mathbf{R}_G$. 

\textbf{Feed-forward Network.} The feed-forward network (FFN) employs the global rotation $\mathbf{R}_G$ alongside local transformations $\mathbf{P}_3$ and $\mathbf{R}_3$. The inputs to the up and gate projections are transformed by the merged weights $\mathbf{R}_G^\top \mathbf{W}_u$ and $\mathbf{R}_G^\top \mathbf{W}_g$, respectively. The local permutation $\mathbf{P}_3$ and rotation $\mathbf{R}_3$ are applied after the activation function; during inference, they will be fused and computed online to smooth the hidden states prior to quantization. Finally, similar to the attention module, the inverse transformations $\mathbf{R}_3^\top$ and $\mathbf{P}_3^\top$ are fused into the down projection weight $\mathbf{W}_d$, and the output is re-projected into the global rotated space via $\mathbf{R}_G$, resulting in the merged weight $\mathbf{R}_3^\top \mathbf{P}_3^\top \mathbf{W}_d \mathbf{R}_G$ without introducing additional inference overhead.

\section{Experiments}

\subsection{Experimental Setup}

\textbf{Evaluation Metrics and Baselines.} We primarily validate the effectiveness of LIRQuant on the LLaMA-2 \citep{touvron2023llama} and LLaMA-3 \citep{grattafiori2024llama} model families. To further examine architectural generalization, we extend our evaluation to Mistral-7B-v0.3 \citep{Jiang2023Mistral7} , Qwen3-8B \citep{yang2025qwen3}, and the Mixture-of-Experts model Mixtral-8$\times$7B \citep{jiang2024mixtral}. Following existing studies \citep{shao2023omniquant, ashkboos2024quarot}, we evaluate the language generation performance using Perplexity (PPL) on the WikiText-2 \citep{merity2016pointer} and C4 \citep{raffel2023exploringlimitstransferlearning} datasets. For common sense reasoning, we select six zero-shot evaluation tasks: ARC-Challenge, ARC-Easy \citep{allenai:arc}, HellaSwag \citep{zellers2019hellaswag}, LAMBADA \citep{paperno-EtAl:2016:P16-1}, PIQA \citep{Bisk2020}, and WinoGrande \citep{sakaguchi2021winogrande}. We compare LIRQuant with main INT4 PTQ methods, including RTN, GPTQ, AWQ, QuIP, SmoothQuant, as well as recent state-of-the-art methods: QuaRot \citep{ashkboos2024quarot}, SpinQuant \citep{liu2025spinquant}, OSTQuant \citep{hu2025ostquant}, FlatQuant \citep{sun2024flatquant}, and DartQuant \citep{shao2026dartquant}.

\textbf{Implementation Details.} Our implementation is based on HuggingFace \citep{DBLP:journals/corr/abs-1910-03771} and PyTorch \citep{DBLP:journals/corr/abs-1912-01703}. We initialize the trainable matrices as diagonal matrices and optimize them using the RiemannAdam optimizer \citep{DBLP:journals/corr/abs-1810-00760} for 200 iterations. The initial learning rate is set to $3 \times 10^{-2}$ with a cosine annealing learning rate decay strategy. Following the standard experimental setup of GPTQ, our calibration set consists of 128 samples from the WikiText-2 dataset, each with a sequence length of 2048 tokens, and the batch size is set to 8. To ensure deployment-friendly inference, we adopt symmetric per-channel quantization for weights and symmetric per-token quantization for activations. For a fair comparison with other baseline methods, we employ Round-to-nearest (RTN) as the activation quantizer and GPTQ as the weight quantizer. Unless otherwise specified, the KV cache is quantized to 4 bits in all weight-activation experiments; that is, the W4A4 setting in this paper always refers to W4A4KV4. Calibration overhead is measured on a single-node 8$\times$NVIDIA H20 (96 GB) platform, and inference efficiency is measured on a single NVIDIA A100-SXM4-80GB.

\subsection{Main Results}
\begin{table*}[t]
\centering
% 1. 缩小基础字号，减少 resizebox 的压缩压力
\small 
% 2. 缩小列间距
\setlength{\tabcolsep}{4pt} 
\renewcommand{\arraystretch}{0.9}
\resizebox{\textwidth}{!}{%
% 3. 彻底删除竖线 |
\begin{tabular}{ll ccc ccc ccc ccc}
\toprule[1.5pt]
\multirow{2}{*}{\makecell{\textbf{Bits}\\\textbf{(W-A-KV)}}} &
\multirow{2}{*}{\textbf{Method}} &
\multicolumn{3}{c}{\textbf{LLaMA-2 7B}} &
\multicolumn{3}{c}{\textbf{LLaMA-2 13B}} &
\multicolumn{3}{c}{\textbf{LLaMA-2 70B}} &
\multicolumn{3}{c}{\textbf{LLaMA-3 8B}} \\
% 4. 使用 \cmidrule 替代竖线来区分不同的模型组，(lr) 表示左右稍微缩进，更美观
\cmidrule(lr){3-5} \cmidrule(lr){6-8} \cmidrule(lr){9-11} \cmidrule(lr){12-14}
& & Wiki $\downarrow$ & C4 $\downarrow$ & Avg $\uparrow$
& Wiki $\downarrow$ & C4 $\downarrow$ & Avg $\uparrow$
& Wiki $\downarrow$ & C4 $\downarrow$ & Avg $\uparrow$
& Wiki $\downarrow$ & C4 $\downarrow$ & Avg $\uparrow$ \\
\midrule

\rowcolor{graybg}
16-16-16 & FP16
& 5.68 & 7.08 & 69.85
& 4.88 & 6.47 & 72.54
& 3.32 & 5.71 & 77.09
& 6.24 & 8.95 & 73.30 \\
\midrule

\multirow{5}{*}{4-16-16}
& RTN
& 5.96 & 7.38 & 65.41
& \textbf{4.98} & 6.59 & 64.56
& 3.87 & 8.91 & 75.13
& 6.82 & 9.72 & 68.92 \\
& GPTQ
& \textbf{5.65} & 7.16 & 66.59
& 5.02 & 6.59 & 70.28
& 4.00 & 7.61 & 76.59
& 6.72 & \textbf{9.55} & 64.65 \\
& AWQ
& 5.69 & 7.20 & 68.01
& 5.01 & 6.61 & 72.08
& 4.10 & 7.35 & 76.53
& 6.83 & 9.76 & 72.08 \\
& QuIP
& 5.87 & 7.34 & -
& 5.05 & 6.69 & -
& - & - & -
& 6.82 & 9.92 & - \\
\rowcolor{oursbg}
& \textbf{LIRQuant (Ours)}
& 5.69 & \textbf{7.11} & \textbf{69.25}
& \textbf{4.98} & \textbf{6.55} & \textbf{72.25}
& \textbf{3.65} & \textbf{6.36} & \textbf{77.38}
& \textbf{6.45} & \textbf{9.55} & \textbf{73.85} \\
\midrule

\multirow{7}{*}{4-4-4}
& RTN
& 5654.5 & 3181.9 & 30.43
& 5636.0 & 4330.9 & 29.93
& 7e4 & 5e4 & 31.42
& 4282.4 & 2936.5 & 31.58 \\
& SmoothQuant
& 9242.4 & 6961.6 & 30.07
& 4731.0 & 3820.3 & 32.63
& 1e3 & 34.61 & 34.01
& 25820.2 & 19069.3 & 30.72 \\
& SpinQuant
& 6.14 & 7.67 & 66.40
& 5.26 & 7.00 & 69.10
& 3.61 & 6.07 & 76.18
& 7.43 & 11.04 & 69.11 \\
& QuaRot
& 6.10 & 8.71 & 66.00
& 5.39 & 7.82 & 69.73
& 3.80 & 6.42 & 76.19
& 10.84 & 18.94 & 65.30 \\
& OSTQuant
& 5.96 & 8.01 & 67.43
& 5.40 & 6.81 & 70.14
& 3.59 & 5.95 & 76.20
& 6.83 & 9.55 & 70.09 \\
& FlatQuant
& \textbf{5.79} & \textbf{7.49} & 67.47
& 5.12 & 6.91 & 71.64
& 3.54 & 5.85 & 76.53
& 7.12 & 10.22 & 71.33 \\
\rowcolor{oursbg}
& \textbf{LIRQuant (Ours)}
& 5.83 & 7.95 & \textbf{67.67}
& \textbf{4.94} & \textbf{6.67} & \textbf{72.05}
& \textbf{3.50} & \textbf{5.81} & \textbf{76.85}
& \textbf{6.81} & \textbf{9.24} & \textbf{71.60} \\
\bottomrule[1.5pt]
\end{tabular}%
}
\caption{Perplexity (PPL) and zero-shot common sense reasoning evaluation of various quantization methods on the LLaMA family under W4A16KV16 and W4A4KV4. For PPL (WikiText-2 and C4), lower is better. For Avg, higher is better. Best results are \textbf{bolded}.}
\label{tab:main_results}
\end{table*}

Table~\ref{tab:main_results} presents the perplexity on WikiText-2 and C4 datasets, alongside the average accuracy on zero-shot common sense reasoning tasks across the LLaMA-2 and LLaMA-3 model families. In W4A16, LIRQuant demonstrates highly competitive performance, matching or outperforming existing state-of-the-art methods like GPTQ and AWQ.

When transitioning to the more challenging W4A4KV4 ultra-low-bit quantization, the advantages of LIRQuant become even more pronounced. For instance, on the LLaMA-2 70B model, LIRQuant achieves a WikiText-2 perplexity of 3.50. This leaves a marginal gap of merely 0.18 compared to the 3.32 perplexity of the full-precision baseline, thereby largely preserving the original generative capacity. Notably, on the newer LLaMA-3 8B model, LIRQuant establishes the optimal results under the W4A4KV4 setting with a perplexity of 6.81 on WikiText-2 and 9.24 on C4. This significantly outperforms recent advanced methods, surpassing SpinQuant by 0.62 and 1.80 points, and FlatQuant by 0.31 and 0.98 points on the two datasets, respectively.

Furthermore, in zero-shot common sense reasoning, LIRQuant consistently delivers the highest average accuracy across all evaluated models under both quantization settings. Detailed results for these downstream tasks are provided in Appendix~\ref{appendix:down_stream}.

\subsection{Generalization to Broader Architectures}

\begin{table}[t]
\centering
\resizebox{\columnwidth}{!}{%
\begin{tabular}{l cc cc cc}
\toprule
\multirow{2}{*}{\textbf{Method}} &
\multicolumn{2}{c}{\textbf{Mistral-7B}} &
\multicolumn{2}{c}{\textbf{Qwen3-8B}} &
\multicolumn{2}{c}{\textbf{Mixtral-8$\times$7B}} \\
\cmidrule(lr){2-3} \cmidrule(lr){4-5} \cmidrule(lr){6-7}
& Wiki $\downarrow$ & Avg $\uparrow$ & Wiki $\downarrow$ & Avg $\uparrow$ & Wiki $\downarrow$ & Avg $\uparrow$ \\
\midrule
SpinQuant  & 5.82 & 70.80 & 7.99 & 69.23 & -- & -- \\
OSTQuant   & 5.79 & 71.10 & 7.77 & 69.70 & OOM & OOM \\
DartQuant  & 5.92 & 70.35 & 8.19 & 68.65 & 5.06 & 75.93 \\
FlatQuant  & 5.76 & 71.38 & 7.44 & 70.57 & 4.27 & 76.79 \\
\rowcolor{oursbg}
\textbf{LIRQuant (Ours)} & \textbf{5.70} & \textbf{71.77} & \textbf{7.32} & \textbf{71.00} & \textbf{4.14} & \textbf{77.27} \\
\bottomrule
\end{tabular}%
}
\caption{WikiText-2 perplexity and average zero-shot accuracy under W4A4KV4 on Mistral-7B-v0.3, Qwen3-8B, and Mixtral-8$\times$7B. OSTQuant results on Mixtral are unavailable due to out-of-memory errors during calibration. Per-task results are in Appendix~\ref{appendix:down_stream}.}
\label{tab:generalization}
\end{table}

To verify that the sensitivity-guided design is not specific to the LLaMA architecture, we further evaluate LIRQuant on Mistral-7B-v0.3, Qwen3-8B, and the MoE model Mixtral-8$\times$7B under W4A4KV4. As shown in Table~\ref{tab:generalization}, LIRQuant ranks first on both WikiText-2 perplexity and the average accuracy of six downstream tasks across all three architectures. On Mixtral-8$\times$7B, LIRQuant achieves a perplexity of 4.14 and an average accuracy of 77.27\%, which are respectively 0.13 lower and 0.48 percentage points higher than the runner-up FlatQuant, while OSTQuant fails to complete calibration due to out-of-memory errors. These results demonstrate that the heterogeneity of channel sensitivity is a general property across modern LLM architectures, including MoE models.

\subsection{Efficiency Analysis}

\begin{table*}[t]
\centering
\small
\setlength{\tabcolsep}{4pt}
\resizebox{\textwidth}{!}{%
\begin{tabular}{l ccc ccc ccc ccc}
\toprule
\multirow{2}{*}{\textbf{Method}} &
\multicolumn{3}{c}{\textbf{LLaMA-3-8B}} &
\multicolumn{3}{c}{\textbf{Mistral-7B-v0.3}} &
\multicolumn{3}{c}{\textbf{Qwen3-8B}} &
\multicolumn{3}{c}{\textbf{Mixtral-8$\times$7B}} \\
\cmidrule(lr){2-4} \cmidrule(lr){5-7} \cmidrule(lr){8-10} \cmidrule(lr){11-13}
& Time & GPU & Host & Time & GPU & Host & Time & GPU & Host & Time & GPU & Host \\
\midrule
SpinQuant  & 9h21m & 26.1 & 16  & 9h19m & 28.6 & 17  & 9h20m & 29.3 & 17  & --    & --   & --  \\
OSTQuant   & 44m   & 46.5 & 421 & 42m   & 33.1 & 443 & 42m   & 60.4 & 419 & --    & OOM  & --  \\
FlatQuant  & 1h13m & 32.4 & 27  & 1h11m & 35.2 & 28  & 1h15m & 38.1 & 30  & 3h12m & 64.5 & 109 \\
DartQuant  & 30m   & 22.7 & 201 & 30m   & 22.1 & 200 & \textbf{30m} & 22.3 & 220 & 1h32m & 16.0 & 304 \\
\rowcolor{oursbg}
\textbf{LIRQuant (Ours)} & \textbf{25m} & \textbf{20.3} & \textbf{16} & \textbf{25m} & \textbf{18.7} & \textbf{15} & \textbf{30m} & \textbf{21.0} & \textbf{16} & \textbf{1h12m} & 35.0 & \textbf{56} \\
\bottomrule
\end{tabular}%
}
\caption{Calibration overhead under W4A4KV4 on 8$\times$NVIDIA H20. Time is the total wall-clock time for calibration (rotation learning + weight quantization); GPU and Host denote per-device peak GPU memory and peak host memory (GB). All baselines use their official implementations. Per-method analysis is provided in Appendix~\ref{appendix:overhead_analysis}.}
\label{tab:overhead}
\end{table*}

\textbf{Calibration Overhead.} Existing orthogonal rotation quantization methods typically require learning or applying dense rotation matrices in the full channel space $\mathbb{R}^{d \times d}$, leading to a quadratic growth in learnable parameters and computational complexity ($\mathcal{O}(d^2)$). In contrast, LIRQuant only learns the transformation matrix $M_{hi}$ within the high-sensitivity subspace $C_{hi}$: under the default 50\% split, the learnable parameters are reduced to one quarter of a dense rotation, which cuts the peak training VRAM from 81.25 GB to 20.31 GB (\textbf{-75\%}) and shortens the end-to-end training time by $2.44\times$ on a single H20 GPU, compared with learning a full-dimensional rotation under the identical training recipe. Table~\ref{tab:overhead} further compares the calibration overhead against four recent rotation-based methods under identical W4A4KV4 settings. LIRQuant retains the same Stiefel-manifold optimizer and end-to-end training paradigm as OSTQuant, but the reduced parameter count directly lowers the per-step backpropagation cost and optimizer state size, and fewer learnable parameters also reduce the overfitting risk and the number of steps required for convergence. Consequently, LIRQuant completes calibration of LLaMA-3-8B in 25 minutes with 20.3 GB peak GPU memory and 16 GB host memory, achieving the best calibration time and host memory on every evaluated model and the lowest GPU memory on all dense models: it avoids SpinQuant's 9-hour float64 re-rotation bottleneck, OSTQuant's teacher-induced GPU memory peaks, and DartQuant's 200 GB-level host memory for activation caching, and it scales to the MoE model Mixtral-8$\times$7B where OSTQuant runs out of memory. A detailed analysis of each baseline's bottleneck is deferred to Appendix~\ref{appendix:overhead_analysis}.

\begin{table}[t]
\centering
\resizebox{\columnwidth}{!}{%
\begin{tabular}{r cccc}
\toprule
\textbf{Seq. Len.} & \textbf{Prefill Speedup} & \textbf{FP16 Prefill} & \textbf{Decode Speedup} & \textbf{Mem. Saving} \\
\midrule
128  & 2.18$\times$ & 31.2 ms   & 0.95$\times$ & 3.55$\times$ \\
256  & 2.20$\times$ & 45.6 ms   & 1.02$\times$ & 3.52$\times$ \\
1024 & 2.25$\times$ & 156.9 ms  & 1.15$\times$ & 3.45$\times$ \\
4096 & 2.29$\times$ & 624.0 ms  & 1.28$\times$ & 3.38$\times$ \\
8192 & 2.32$\times$ & 1320.5 ms & 1.40$\times$ & 3.30$\times$ \\
\bottomrule
\end{tabular}%
}
\caption{End-to-end inference efficiency of LIRQuant on LLaMA-3-8B (batch size 4, NVIDIA A100-80GB) relative to the FP16 baseline. Results on Mistral-7B and Qwen3-8B, and the batch-size scaling study, are in Appendix~\ref{appendix:inference_details}.}
\label{tab:inference}
\end{table}

\textbf{Inference Efficiency.} During inference, only two transformations remain online: $\mathbf{R}_1$ applied to Q/K after RoPE and $\mathbf{R}_3$ before the down projection, while $\mathbf{R}_G$ and $\mathbf{R}_2$ are absorbed offline into the weights. Both online transformations preserve the block-diagonal structure $\mathrm{blkdiag}(M_{hi}, H_{lo})$, so the dense sub-block only operates on the high-sensitivity dimensions and the remainder uses the fast Hadamard transform, reducing computation and memory traffic. We fuse the block-diagonal transform, per-token reduction, symmetric INT4 quantization, and packing into a single Triton kernel, and adopt CUDA Graph for fixed-shape decoding. As shown in Table~\ref{tab:inference}, on LLaMA-3-8B LIRQuant achieves $2.18\times$--$2.32\times$ prefill speedup and up to $3.55\times$ peak memory saving over the FP16 baseline; the decode speedup grows with sequence length and batch size, reaching $1.82\times$ at batch size 64. Detailed kernel designs and full results are provided in Appendix~\ref{appendix:inference_details}.

\subsection{Ablation Study}

\begin{table}[t]
\centering
\resizebox{\columnwidth}{!}{%
\begin{tabular}{l ccccccccccc}
\toprule
\textbf{Ratio (\%)} & 0 & 10 & 20 & 30 & 40 & 50 & 60 & 70 & 80 & 90 & 100 \\
\midrule
\textbf{PPL} & 11.59 & 13.28 & 8.92 & 7.55 & 7.32 & \textbf{6.78} & 6.82 & 6.88 & 6.85 & 6.92 & 8.01 \\
\bottomrule
\end{tabular}%
}
\caption{WikiText-2 PPL of LLaMA-3-8B (W4A4KV4) as a function of the high-sensitivity ratio $r$. $r=0\%$ corresponds to the full-Hadamard configuration and $r=100\%$ to a fully learnable dense rotation.}
\label{tab:ratio_sweep}
\end{table}

\textbf{High-Sensitivity Ratio.} Table~\ref{tab:ratio_sweep} sweeps the high-sensitivity ratio $r$ from 0\% (full Hadamard) to 100\% (fully learnable dense rotation). The sweep is conducted with an independently re-sampled calibration set, so the absolute PPL values may differ slightly from those in Table~\ref{tab:main_results}. The PPL exhibits a rise--fall--plateau pattern. At $r=10\%$, a diffusion loss peak emerges: the block-diagonal structure eliminates cross-block energy exchange, so outliers concentrated in a tiny learnable block can no longer be dispersed over the full channel space, temporarily degrading quantization precision. As $r$ increases, the learnable rotation quickly compensates for this loss and the PPL drops steeply, reaching its best value at $r=50\%$. Beyond this point, the PPL merely oscillates within a narrow band without consistent improvement, while the learnable parameters and the associated overfitting risk grow quadratically with $r$. We therefore adopt the 50/50 split as the principled default configuration of LIRQuant. The construction of $\mathbf{H}_\text{lo}$ for arbitrary dimensions is described in Appendix~\ref{appendix:valid_decompositions}, and the comparison against alternative channel selection metrics over the full sweep is provided in Appendix~\ref{appendix:metric_sweep}.

\textbf{Sensitivity Metric.} We compare our noise-weighted functional sensitivity against six common channel importance metrics (maxabs, L2 norm, L1 norm as in AWQ, the 99.9th percentile, kurtosis, and random selection) across the full ratio sweep in Appendix~\ref{appendix:metric_sweep}. Our metric yields the best PPL at every ratio in the practically relevant range (20\%--70\%), including the optimal 50\% split. Among the alternatives, only maxabs approaches our performance owing to its fidelity to extreme outliers, whereas magnitude-accumulation metrics (L1/L2) are diluted by moderate values, and scale-invariant kurtosis discards magnitude information entirely. The additional cost of computing our metric is on the order of minutes, which is negligible relative to the resulting accuracy gains.

\begin{figure}[t]
    \centering
    \includegraphics[width=\linewidth]{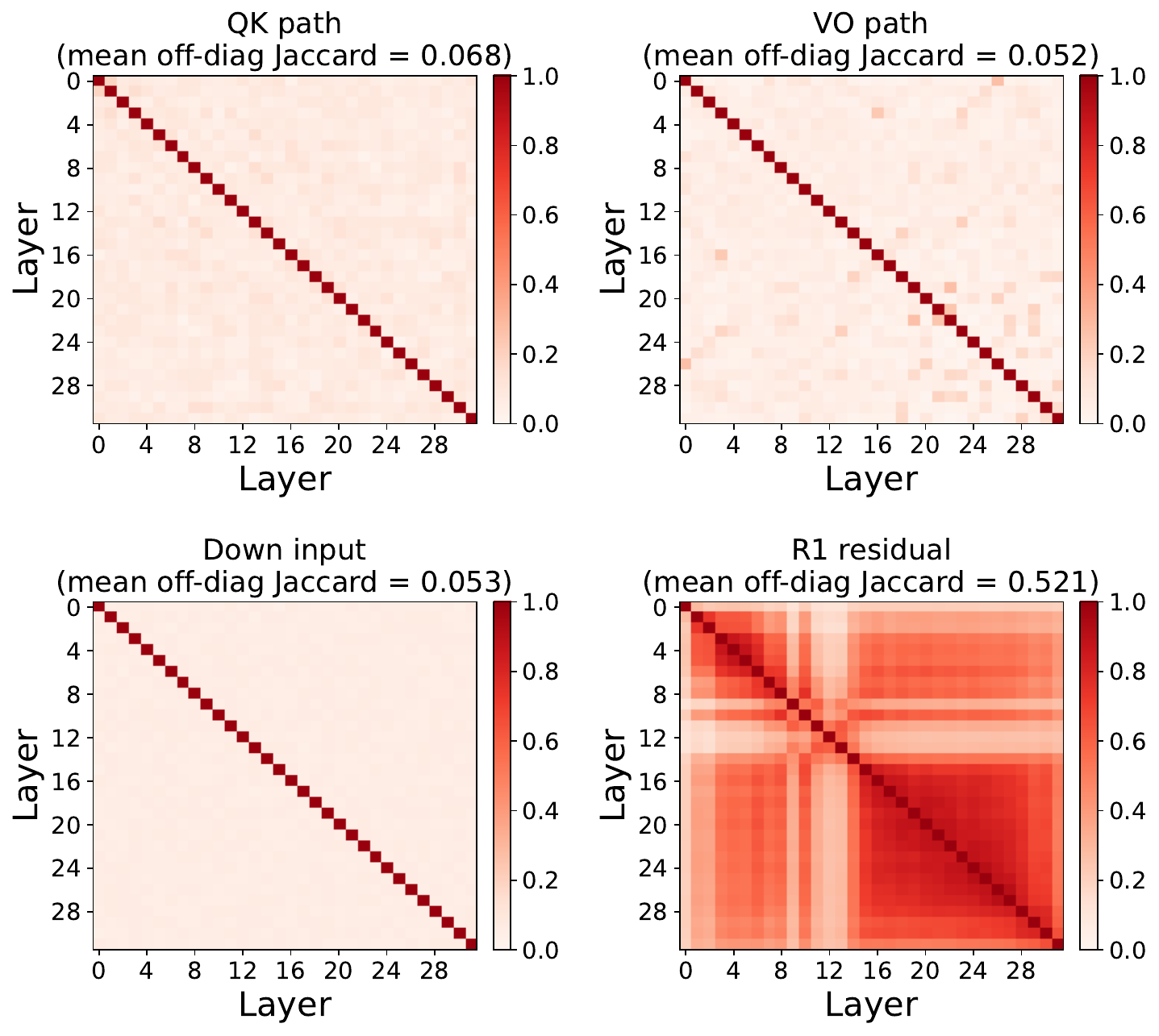}
    \caption{Heatmap of pairwise Jaccard similarity between the sets of top-10\% sensitive channel positions from different layers.}
    \label{fig:jaccard}
\end{figure}

\textbf{Cross-Layer Independence Analysis.} As shown in Figure~\ref{fig:jaccard}, the four paths exhibit distinctly different inter-layer structures. The average off-diagonal Jaccard coefficient for the residual stream (labeled R1 in the figure) is 0.521, indicating a significant overlap of high-sensitivity channels across different layers. This aligns with the global semantic consistency of the residual stream and provides direct justification for sharing a single global rotation matrix $\mathbf{R}_G$ across layers. In contrast, the inter-layer Jaccard coefficients for the QK path, VO path, and Down input are all below 0.07, and their heatmaps display an almost purely diagonal sparse structure, meaning the high-sensitivity channels across different layers have nearly zero overlap. This result demonstrates that the sensitive channels at local rotation positions possess strong layer specificity. Any attempt to share local rotation matrices across layers would lead to a significant loss in coverage, thereby providing empirical support for our Stage-2 design of independently training $\mathbf{R}_1^{(l)}$, $\mathbf{R}_2^{(l)}$, and $\mathbf{R}_3^{(l)}$ for each layer.

\begin{figure}[t]
    \centering
    \includegraphics[width=\linewidth]{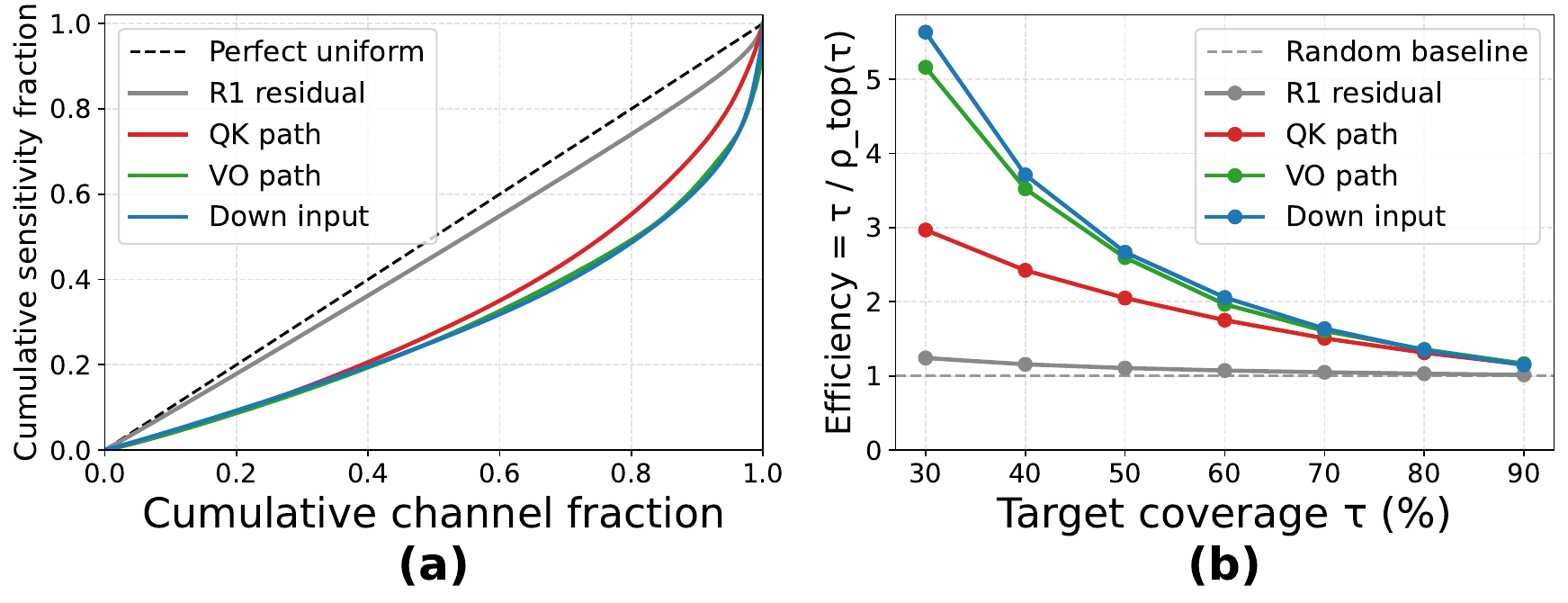}
    \caption{(a) Lorenz curves of channel sensitivity. (b) Channel reduction factor of sensitivity-guided selection compared with random selection under a target coverage ratio $\tau$.}
    \label{fig:lorenz}
\end{figure}

\textbf{Intra-Layer Concentration Analysis.} As illustrated in Figure~\ref{fig:lorenz}, the Lorenz curves reveal significant differences in the intra-layer distributions across various paths. The diagonal dashed line represents the theoretical baseline of a perfectly uniform distribution. The curve for the QK path deviates most significantly from the diagonal (Gini = 0.63), whereas the Down input and the residual stream remain close to the diagonal with Gini coefficients of 0.24 and 0.10, respectively. This discrepancy is reflected more intuitively as an efficiency multiplier in Figure~\ref{fig:lorenz}(b): at $\tau=50\%$, the sensitivity-guided selection efficiency for the QK path reaches up to 7.5$\times$---meaning it requires only $1/7.5$ of the channels compared to random selection to cover the same amount of sensitivity mass. The VO path achieves 2.8$\times$, the Down input 1.5$\times$, and the residual stream merely 1.1$\times$, which is almost indistinguishable from the random baseline. Cross-layer consistency determines whether global or layer-wise rotations should be adopted, while intra-layer concentration dictates the gain efficiency brought by coverage-based channel selection. Considering both factors, we apply a dense global rotation to the residual stream, and employ sensitivity-based local rotations for the remaining paths.

\section{Conclusion}
In this paper, we propose LIRQuant, an efficient Locally Isolated Rotation Quantization framework. By strategically isolating high-sensitivity channels into a learnable subspace while applying fixed Hadamard transforms to the remaining low-sensitivity channels, our method successfully balances channel mixing quality and parameter efficiency. Under the challenging W4A4KV4 setting, LIRQuant achieves state-of-the-art accuracy across the LLaMA, Mistral, Qwen3, and Mixtral families, completes calibration of LLaMA-3-8B in 25 minutes with 20.3 GB peak GPU memory---the lowest overhead among recent rotation-based methods---and delivers up to $2.3\times$ prefill speedup and $3.5\times$ memory saving at inference, making the efficient optimization and deployment of massive language models on resource-constrained devices (e.g., a single GPU) highly feasible.

\section*{Limitations}
LIRQuant determines the channel partition statically from a small calibration set: the partition is shared across all inputs, and the 50\% high-sensitivity ratio is applied uniformly to every insertion point, whereas a per-layer or input-adaptive ratio may further improve the accuracy-efficiency trade-off. In addition, our study focuses on INT4 weight-activation-KV quantization; extending the sensitivity-guided isolation to more aggressive settings (e.g., W3A3 or low-bit floating-point formats) and to larger-scale MoE models is left for future work.

\bibliography{latex/custom}

\clearpage
\appendix
\section{Sensitivity Derivation}\label{sec:sensitivity_derivation}
In this section, we provide a detailed derivation of the noise-weighted local functional sensitivity metric introduced in Section~\ref{sensitivity}. Our objective is to quantify the expected distortion on the local target output $y_p = f_p(z_p)$ caused by introducing a quantization perturbation $\epsilon_p$ to the activation $z_p$ at insertion point $p$.

Given a calibration set $\mathcal{C}$, the local functional distortion is defined as the expected distance between the outputs before and after the perturbation:
\begin{equation}
\mathcal{D}_p = \mathbb{E}_{\mathcal{C}} \bigl[ D_p\bigl(f_p(z_p+\epsilon_p), f_p(z_p)\bigr) \bigr],
\end{equation}
where $D_p$ is a distance metric (e.g., Mean Squared Error). To analyze the impact of a small perturbation $\epsilon_p$, we perform a second-order Taylor expansion of the distance function around the unperturbed activation $z_p$. 

Let $L(z_p) = D_p\bigl(f_p(z_p), f_p(z_p)\bigr)$. Trivially, when there is no perturbation, the distance is zero, meaning the zeroth-order term is $L(z_p) = 0$. Furthermore, since $z_p$ is the global minimum of this distance function with respect to itself, the first-order derivative (gradient) is also zero, i.e., $\nabla L(z_p) = \mathbf{0}$. Therefore, the dominant term in the Taylor expansion is the second-order term:
\begin{equation}
D_p\bigl(f_p(z_p+\epsilon_p), f_p(z_p)\bigr) \approx \frac{1}{2} \epsilon_p^\top H_p \epsilon_p,
\end{equation}
where $H_p = \nabla^2_{z_p} D_p\bigl(f_p(z_p), f_p(z_p)\bigr)$ is the Hessian matrix of the distance metric with respect to the input activation $z_p$.

Next, we take the expectation over both the calibration data distribution and the quantization noise. By applying the cyclic permutation property of the trace operator ($\text{Tr}$), we can rewrite the expected distortion as:
% \begin{equation}
% \mathcal{D}_p \approx \mathbb{E}_{\mathcal{C}, \epsilon_p} \left[ \frac{1}{2} \text{Tr}\left(\epsilon_p^\top H_p \epsilon_p\right) \right] = \mathbb{E}_{\mathcal{C}} \left[ \frac{1}{2} \text{Tr}\left(H_p \cdot \mathbb{E}_{\epsilon_p}[\epsilon_p \epsilon_p^\top]\right) \right].
% \end{equation}
\begin{equation}
\begin{split}
\mathcal{D}_p &\approx \mathbb{E}_{\mathcal{C}, \epsilon_p} \left[ \frac{1}{2} \text{Tr}\left(\epsilon_p^\top H_p \epsilon_p\right) \right] \\
&= \mathbb{E}_{\mathcal{C}} \left[ \frac{1}{2} \text{Tr}\left(H_p \cdot \mathbb{E}_{\epsilon_p}[\epsilon_p \epsilon_p^\top]\right) \right].
\end{split}
\end{equation}

In standard quantization theory, it is commonly assumed that the quantization noise across different channels is zero-mean and mutually independent. Consequently, the covariance matrix of the quantization noise $\mathbb{E}_{\epsilon_p}[\epsilon_p \epsilon_p^\top]$ is a diagonal matrix, where the diagonal elements represent the noise variance $\sigma_{p,j}^2$ of each channel:
\begin{equation}
\mathbb{E}_{\epsilon_p}[\epsilon_p \epsilon_p^\top] = \text{diag}(\sigma_{p,1}^2, \sigma_{p,2}^2, \dots, \sigma_{p,d}^2).
\end{equation}

We define the functional sensitivity matrix as half of the expected Hessian: $G_p = \frac{1}{2} \mathbb{E}_{\mathcal{C}}[H_p]$. Because the noise covariance matrix is diagonal, the trace operation eliminates all off-diagonal elements of $G_p$. The total distortion $\mathcal{D}_p$ can thus be diagonalized and decomposed into a linear sum of independent contributions from each channel:
\begin{equation}
\begin{split}
\mathcal{D}_p &\approx \text{Tr}\left( G_p \cdot \text{diag}(\sigma_{p,1}^2, \dots, \sigma_{p,d}^2) \right) \\
&= \sum_{j=1}^{d} G_{p,jj} \sigma_{p,j}^2.
\end{split}
\end{equation}

From this, we naturally derive the sensitivity definition for the $j$-th channel:
\begin{equation}
s_{p,j} = G_{p,jj} \sigma_{p,j}^2.
\end{equation}
This formulation decouples the \textit{functional sensitivity} ($G_{p,jj}$, the diagonal element of the expected Hessian, reflecting the model's structural reliance on the channel) from the \textit{quantization noise intensity} ($\sigma_{p,j}^2$, reflecting the information loss incurred during quantization).

\textbf{Extension to Weight Quantization.} When the quantization of the weight matrix $W_p$ is also considered, the local output becomes $y_p = f_p(z_p, W_p)$, and the perturbation includes both activation noise $\epsilon_p$ and weight noise $\delta_p$. Assuming that the activation quantization noise and weight quantization noise are independent (i.e., $\mathbb{E}[\epsilon_p \delta_p^\top] = \mathbf{0}$), the expectation of the cross-terms in the second-order Taylor expansion evaluates to zero. Therefore, the total distortion can be approximated as the linear superposition of the activation distortion and the weight distortion, yielding the additive approximation used in our framework:
\begin{equation}
s_{p,j} \approx s_{p,j}^{act} + s_{p,j}^{w}.
\end{equation}

\section{Long-tail Effect in Local Paths}
\label{appendix:long_tail}
\begin{figure}[htbp]
    \centering
    \includegraphics[width=\linewidth]{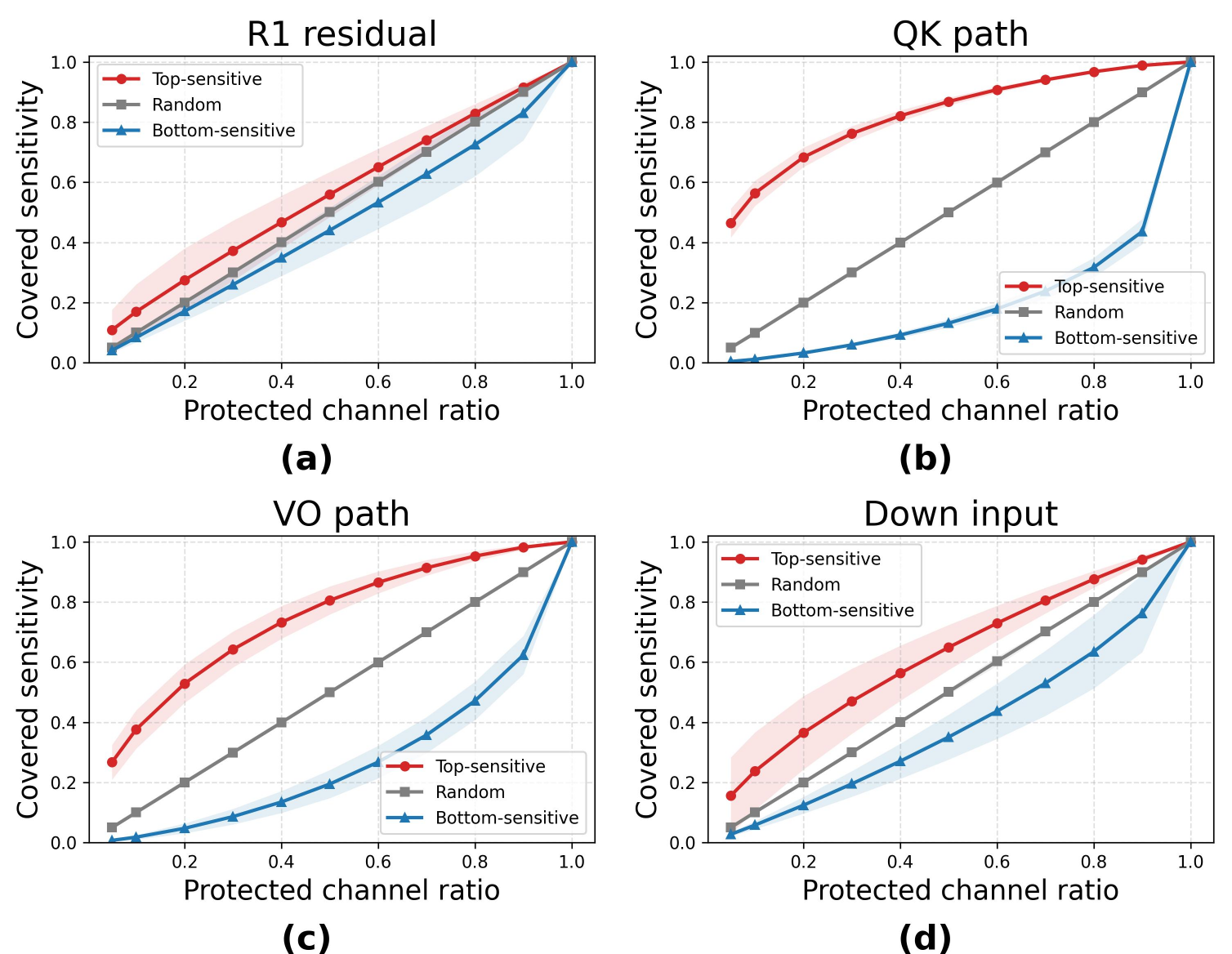}
    \caption{Lorenz curves of intra-layer sensitivity distribution.}
    \label{fig:sanity_combined}
\end{figure}
As shown in Figure~\ref{fig:sanity_combined}(b) and (c), local paths such as the QK and VO paths exhibit a severe long-tail distribution, drastically deviating from the uniform diagonal. For example, the top 20\% of channels in the QK path dominate approximately 70\% of the total sensitivity. This extreme concentration indicates that most of the quantization distortion is caused by a tiny fraction of critical channels, providing direct empirical support for our design of allocating expressive learnable rotations exclusively to these high-sensitivity channels.

\section{Hadamard Transforms for Non-Power-of-Two Dimensions}
\label{appendix:valid_decompositions}

The fast Walsh--Hadamard transform (FWHT) is only defined for power-of-two dimensions, whereas the low-sensitivity dimension $d_{\mathrm{lo}}$ is in general arbitrary. Following the implementation of QuaRot, we factorize the dimension as
\begin{equation}
    d_{\mathrm{lo}} = K \cdot 2^{m},
\end{equation}
where $K$ is the size of a known Hadamard matrix construction, $K \in \{1, 12, 20, 28, 36, 40, 52, 60, 108, 140, 156, 172\}$, and build the transform as a Kronecker product:
\begin{equation}
    \mathbf{H}_{\mathrm{lo}} = \mathbf{H}_{K} \otimes \mathbf{H}_{2^{m}}.
\end{equation}
In computation, the input is reshaped into $K$ groups of size $2^{m}$, transformed by the FWHT along the $2^{m}$ dimension, and then multiplied by the small dense matrix $\mathbf{H}_{K}$, with a total cost of $\mathcal{O}(d_{\mathrm{lo}} \log 2^{m})$ plus one $K \times K$ matrix product. Since the Kronecker product of two Hadamard matrices is itself a Hadamard matrix, $\mathbf{H}_{\mathrm{lo}}$ is a complete orthogonal transform over all $d_{\mathrm{lo}}$ channels, i.e., the outlier energy is mixed across the entire low-sensitivity subspace rather than within isolated fragments.

When $d_{\mathrm{lo}}$ is a power of two, $K=1$ and the transform degenerates to a single FWHT without any additional matrix product, which is the most efficient case. Under our default 50\% split on LLaMA-3-8B, the hidden-dimension paths use $d_{\mathrm{lo}} = 2048$ ($K=1$), and the down-projection input uses $d_{\mathrm{lo}} = 7168 = 28 \times 256$ ($K=28$), matching the fused kernels described in Appendix~\ref{appendix:inference_details}. The set of valid $d_{\mathrm{lo}}$ values is therefore $\{K \cdot 2^{m}\}$ over the supported $K$, which forms a dense grid over the sweep range in Table~\ref{tab:ratio_sweep}; ratios are rounded to the nearest valid dimension.

\section{Detailed Results on Downstream Tasks}
\label{appendix:down_stream}
As shown in Table~\ref{tab:downstream}, LIRQuant consistently outperforms prior SOTA methods across all evaluated models and quantization configurations. Under the W4A16 setting, LIRQuant comprehensively surpasses strong baselines such as GPTQ and AWQ, minimizing the performance gap with full-precision (FP16) models in zero-shot reasoning. In the highly challenging W4A4KV4 ultra-low-bit setting, traditional smoothing methods (e.g., SmoothQuant) suffer from severe performance collapse on specific tasks (e.g., LAMBADA), whereas our approach maintains a substantial lead. Particularly on the notoriously difficult-to-quantize LLaMA-3-8B model, LIRQuant achieves the highest average accuracy of 71.60\%, directly surpassing the latest advanced methods including FlatQuant (71.33\%) and SpinQuant (69.11\%). Achieving this optimal accuracy firmly validates the effectiveness of our core design.
\begin{table*}[htbp]
\centering
\resizebox{\textwidth}{!}{
\begin{tabular}{c | c | l | c c c c c c | c}
\toprule[1.5pt]
\textbf{Model} & \textbf{\#Bits (W-A-KV)} & \textbf{Method} & \textbf{ARC-c} & \textbf{ARC-e} & \textbf{HellaS.} & \textbf{Lam.} & \textbf{PIQA} & \textbf{WinoG.} & \textbf{Avg.} \\
\midrule
\multirow{11}{*}{LLaMA-2-7B} 
& \multirow{4}{*}{4-16-16} 
& RTN & 42.15 & 67.59 & 72.34 & 67.18 & 76.50 & 66.69 & 65.41 \\
& & GPTQ & 42.49 & 69.53 & 73.83 & 67.61 & 77.64 & 68.43 & 66.59 \\
& & AWQ & 44.11 & 70.75 & 74.98 & 70.68 & 78.13 & 69.38 & 68.01 \\
& & \textbf{LIRQuant (Ours)} & \textbf{44.80} & \textbf{73.60} & \textbf{75.30} & \textbf{73.90} & \textbf{79.10} & \textbf{68.80} & \textbf{69.25} \\
\cline{2-10}
& \multirow{7}{*}{4-4-4} 
& RTN & 27.22 & 27.06 & 27.34 & 0.93 & 49.51 & 50.51 & 30.43 \\
& & SmoothQuant & 26.37 & 25.63 & 27.05 & 1.11 & 51.90 & 34.49 & 30.07 \\
& & QuaRot & 41.43 & 69.32 & 72.50 & 70.66 & 77.42 & 64.64 & 66.00 \\
& & SpinQuant & 40.44 & 71.08 & 73.51 & 70.66 & 76.88 & 65.82 & 66.40 \\
& & OSTQuant & 42.92 & 72.56 & 73.14 & 71.76 & 77.42 & 66.77 & 67.43 \\
& & FlatQuant & 43.00 & 71.21 & 73.31 & 72.06 & 77.53 & 67.72 & 67.47 \\
& & \textbf{LIRQuant (Ours)} & \textbf{43.20} & \textbf{71.50} & \textbf{73.50} & \textbf{72.20} & \textbf{77.70} & \textbf{67.90} & \textbf{67.67} \\
\midrule
\multirow{11}{*}{LLaMA-2-13B} 
& \multirow{4}{*}{4-16-16} 
& RTN & 42.92 & 66.54 & 66.62 & 68.99 & 76.93 & 65.35 & 64.56 \\
& & GPTQ & 49.63 & 73.95 & 73.77 & 73.20 & 78.51 & 70.64 & 70.28 \\
& & AWQ & 48.63 & 78.16 & 78.48 & 75.20 & 79.54 & 72.45 & 72.08 \\
& & \textbf{LIRQuant (Ours)} & \textbf{49.00} & \textbf{76.50} & \textbf{78.50} & \textbf{76.80} & \textbf{80.50} & \textbf{72.20} & \textbf{72.25} \\
\cline{2-10}
& \multirow{7}{*}{4-4-4} 
& RTN & 27.82 & 26.52 & 26.27 & 0.02 & 49.78 & 49.17 & 29.93 \\
& & SmoothQuant & 24.49 & 33.00 & 30.70 & 2.70 & 53.81 & 51.07 & 32.63 \\
& & QuaRot & 47.27 & 73.91 & 75.33 & 73.53 & 79.27 & 69.06 & 69.73 \\
& & SpinQuant & 46.67 & 74.49 & 75.22 & 72.19 & 78.29 & 67.72 & 69.10 \\
& & OSTQuant & 47.10 & 75.21 & 76.71 & 75.14 & 78.67 & 68.03 & 70.14 \\
& & FlatQuant & 48.38 & 76.94 & 77.88 & 76.40 & 79.65 & 70.56 & 71.64 \\
& & \textbf{LIRQuant (Ours)} & \textbf{48.80} & \textbf{77.30} & \textbf{78.20} & \textbf{76.90} & \textbf{80.10} & \textbf{71.00} & \textbf{72.05} \\
\midrule
\multirow{11}{*}{LLaMA-3-8B} 
& \multirow{4}{*}{4-16-16} 
& RTN & 48.98 & 73.23 & 75.90 & 63.85 & 78.40 & 73.16 & 68.92 \\
& & GPTQ & 49.74 & 72.52 & 68.34 & 46.69 & 78.78 & 71.82 & 64.65 \\
& & AWQ & 52.22 & 76.68 & 77.51 & 74.81 & 79.60 & 71.67 & 72.08 \\
& & \textbf{LIRQuant (Ours)} & \textbf{53.10} & \textbf{80.10} & \textbf{78.10} & \textbf{76.70} & \textbf{81.00} & \textbf{74.10} & \textbf{73.85} \\
\cline{2-10}
& \multirow{7}{*}{4-4-4} 
& RTN & 23.72 & 30.56 & 29.83 & 2.70 & 52.45 & 50.20 & 31.58 \\
& & SmoothQuant & 23.55 & 28.96 & 28.90 & 1.44 & 51.09 & 50.36 & 30.72 \\
& & QuaRot & 42.83 & 67.42 & 72.66 & 66.93 & 75.73 & 66.22 & 65.30 \\
& & SpinQuant & 46.33 & 73.57 & 75.43 & 71.40 & 79.16 & 68.75 & 69.11 \\
& & OSTQuant & 49.32 & 76.73 & 76.01 & 70.77 & 78.51 & 69.22 & 70.09 \\
& & FlatQuant & 50.51 & 75.88 & 76.49 & 73.20 & 79.00 & 72.93 & 71.33 \\
& & \textbf{LIRQuant (Ours)} & \textbf{50.80} & \textbf{76.10} & \textbf{76.70} & \textbf{73.50} & \textbf{79.30} & \textbf{73.20} & \textbf{71.60} \\
\midrule
\multirow{11}{*}{LLaMA-2-70B} 
& \multirow{4}{*}{4-16-16} 
& RTN & 55.80 & 79.29 & 81.78 & 75.51 & 81.94 & 76.48 & 75.13 \\
& & GPTQ & 56.91 & 80.81 & 82.47 & 79.06 & 82.75 & 77.51 & 76.59 \\
& & AWQ & 56.67 & 80.54 & 82.54 & 78.83 & 82.97 & 77.62 & 76.53 \\
& & \textbf{LIRQuant (Ours)} & \textbf{57.60} & \textbf{81.60} & \textbf{84.10} & \textbf{80.00} & \textbf{82.90} & \textbf{78.10} & \textbf{77.38} \\
\cline{2-10}
& \multirow{7}{*}{4-4-4} 
& RTN & 30.38 & 27.74 & 26.12 & 0.02 & 51.74 & 52.49 & 31.42 \\
& & SmoothQuant & 24.15 & 33.88 & 31.75 & 7.14 & 54.95 & 52.17 & 34.01 \\
& & QuaRot & 56.48 & 80.56 & 81.93 & 79.16 & 82.21 & 76.80 & 76.19 \\
& & SpinQuant & 56.31 & 80.64 & 82.36 & 79.41 & 82.21 & 76.16 & 76.18 \\
& & OSTQuant & 56.58 & 80.17 & 82.49 & 78.72 & 82.76 & 76.49 & 76.20 \\
& & FlatQuant & 56.40 & 80.09 & 82.91 & 80.01 & 82.92 & 76.87 & 76.53 \\
& & \textbf{LIRQuant (Ours)} & \textbf{56.70} & \textbf{80.40} & \textbf{83.20} & \textbf{80.40} & \textbf{83.20} & \textbf{77.20} & \textbf{76.85} \\
\bottomrule[1.5pt]
\end{tabular}%
}
\caption{Detailed zero-shot accuracy on downstream tasks for LLaMA-2 and LLaMA-3 families under W4A16KV16 and W4A4KV4 quantization.}
\label{tab:downstream}
\end{table*}

Table~\ref{tab:downstream_new} reports the per-task zero-shot accuracy on Mistral-7B-v0.3, Qwen3-8B, and Mixtral-8$\times$7B under W4A4KV4, complementing the summary in Table~\ref{tab:generalization}.

\begin{table*}[htbp]
\centering
\resizebox{\textwidth}{!}{
\begin{tabular}{c | l | c | c c c c c c | c}
\toprule[1.5pt]
\textbf{Model} & \textbf{Method} & \textbf{Wiki} $\downarrow$ & \textbf{ARC-c} & \textbf{ARC-e} & \textbf{HellaS.} & \textbf{Lam.} & \textbf{PIQA} & \textbf{WinoG.} & \textbf{Avg.} \\
\midrule
\multirow{5}{*}{Mistral-7B-v0.3}
& SpinQuant & 5.82 & 46.9 & 77.2 & 78.9 & 73.2 & 78.9 & 69.7 & 70.80 \\
& OSTQuant & 5.79 & 47.2 & 77.4 & 79.1 & 73.6 & 79.2 & 70.1 & 71.10 \\
& DartQuant & 5.92 & 46.4 & 76.7 & 78.5 & 72.8 & 78.5 & 69.2 & 70.35 \\
& FlatQuant & 5.76 & 47.5 & 77.7 & 79.3 & 73.9 & 79.5 & 70.4 & 71.38 \\
& \textbf{LIRQuant (Ours)} & \textbf{5.70} & \textbf{47.8} & \textbf{78.1} & \textbf{79.7} & \textbf{74.3} & \textbf{79.8} & \textbf{70.9} & \textbf{71.77} \\
\midrule
\multirow{5}{*}{Qwen3-8B}
& SpinQuant & 7.99 & 52.8 & 79.5 & 74.1 & 64.2 & 76.6 & 68.2 & 69.23 \\
& OSTQuant & 7.77 & 53.3 & 80.0 & 74.6 & 64.7 & 77.1 & 68.8 & 69.70 \\
& DartQuant & 8.19 & 52.3 & 78.9 & 73.5 & 63.6 & 76.1 & 67.5 & 68.65 \\
& FlatQuant & 7.44 & 54.1 & 80.8 & 75.4 & 65.6 & 77.8 & 69.7 & 70.57 \\
& \textbf{LIRQuant (Ours)} & \textbf{7.32} & \textbf{54.4} & \textbf{81.2} & \textbf{75.8} & \textbf{66.1} & \textbf{78.2} & \textbf{70.3} & \textbf{71.00} \\
\midrule
\multirow{3}{*}{Mixtral-8$\times$7B}
& DartQuant & 5.06 & 58.60 & 82.45 & 81.78 & 77.32 & 81.55 & 73.85 & 75.93 \\
& FlatQuant & 4.27 & 59.30 & 83.21 & 82.92 & 78.13 & 82.26 & 74.90 & 76.79 \\
& \textbf{LIRQuant (Ours)} & \textbf{4.14} & \textbf{59.54} & \textbf{83.65} & \textbf{83.48} & \textbf{78.55} & \textbf{82.84} & \textbf{75.62} & \textbf{77.27} \\
\bottomrule[1.5pt]
\end{tabular}%
}
\caption{Detailed zero-shot accuracy and WikiText-2 perplexity on Mistral-7B-v0.3, Qwen3-8B, and Mixtral-8$\times$7B under W4A4KV4. SpinQuant and OSTQuant results on Mixtral are unavailable due to out-of-memory errors during calibration.}
\label{tab:downstream_new}
\end{table*}

%----------------------------------------------------------------
% 新增 Appendix E
%----------------------------------------------------------------
\section{Detailed Analysis of Calibration Overhead}
\label{appendix:overhead_analysis}

This section analyzes the calibration bottlenecks of each baseline in Table~\ref{tab:overhead}. All experiments are conducted on a single-node 8$\times$NVIDIA H20 (96 GB per device) platform, with all competing methods using their official open-source implementations and quantization uniformly configured as W4A4KV4. Both GPU and host memory are reported per device to avoid incomparability arising from data-parallel replicas.

\textbf{SpinQuant} learns rotations via end-to-end cross-entropy optimization with Cayley-SGD on the Stiefel manifold. However, its official implementation performs on-the-fly weight re-rotation in float64 for every linear layer with per-sample training (batch size 1), resulting in a calibration time as high as 9 hours, the slowest among all methods. Its memory advantage (26 GB) stems from frozen weights, gradient checkpointing, and DDP with full per-device replicas, requiring neither a teacher model nor activation caching.

\textbf{OSTQuant} compresses the calibration time to about 44 minutes via top-K KL distillation and per-head orthogonal transforms. However, its end-to-end training superimposes a per-step full-precision teacher forward pass and full-weight float32 re-rotation, yielding the highest per-device peak GPU memory (33--60 GB). Its $\sim$420 GB host memory is a transient spike during initialization (float64 LayerNorm folding), with the steady state falling back to $\sim$16 GB. On Mixtral, the end-to-end framework must simultaneously hold the quantized model, the teacher, and the re-rotation buffers, exceeding the total GPU memory and causing OOM.

\textbf{FlatQuant} adopts layer-wise Kronecker affine reconstruction with strictly sequential training across layers, precluding multi-device parallelism and taking about 1h13m. On Mixtral, the 8-expert activation cache drives the host memory to 109 GB and the wall-clock time to 3h12m.

\textbf{DartQuant} achieves fast calibration by trading memory for computation: it decouples rotation learning from the model and trains on cached activations. The calibration time on dense models is only about 30 minutes, but it requires $\sim$200 GB of sustained host memory (as opposed to OSTQuant's transient spike), and the global rotation uses only a single device. To train Mixtral on our cluster, we modified its caching mechanism, yet the host memory still reached 304 GB.

\textbf{LIRQuant} adopts the same Stiefel optimizer as OSTQuant and retains end-to-end training, but reduces the number of learnable parameters to one quarter of a dense rotation. This directly cuts the per-step backpropagation computation and optimizer state size, effectively controlling GPU memory overhead; fewer learnable parameters also naturally lower the risk of overfitting and reduce the number of steps required for convergence. In our implementation, we further apply FSDP weight sharding and use float32 rotations to circumvent the float64 DGEMM bottleneck and host memory spikes. As a result, LIRQuant achieves the best calibration time and host memory on every evaluated model; on Mixtral, DartQuant attains a lower GPU footprint only because it trains a single global rotation on cached activations, at the cost of 304 GB host memory.

%----------------------------------------------------------------
% 新增 Appendix F
%----------------------------------------------------------------
\section{Inference Efficiency Details}
\label{appendix:inference_details}

All inference experiments are conducted on a single NVIDIA A100-SXM4-80GB, with the FP16 unquantized model as the baseline. We evaluate end-to-end performance through two experiments: Experiment 1 fixes the batch size at 4 and sweeps the sequence length (128--8192) to examine prefill/decode speedup and memory savings as a function of sequence length; Experiment 2 fixes the prefill length at 2048 and sweeps the batch size (1--64) to examine decode speedup scalability with concurrency.

\subsection{Kernel Design}

LIRQuant retains only two online rotations: $\mathbf{R}_1$ on Q/K after RoPE and $\mathbf{R}_3$ before the down projection; $\mathbf{R}_G$ and $\mathbf{R}_2$ ($V \to W_O$) are absorbed offline into weights. We adopt the block-diagonal structure $\mathrm{blkdiag}(M_{hi}, H_{lo})$---only high-sensitivity dimensions use the learnable dense sub-block $M_{hi}$, while the rest use the fixed Hadamard transform $H_{lo}$---reducing computation and memory traffic while preserving expressiveness. The kernels are designed at three levels.

\textbf{Operator Fusion.} Both sub-blocks in $\mathbf{R}_3$ reduce to Kronecker structures ($y_d = L^\top (x_d R_r)$, $H_{7168} = H_{28} \otimes H_{256}$). We fuse the block-diagonal transform, per-token reduction, symmetric INT4 quantization, and dual-nibble packing into a single Triton program: the dense and Hadamard inputs are loaded from global memory, transformed on-chip with all intermediates in registers, jointly reduced, quantized and packed online, and written back in a single pass together with the per-token scales. The output is directly consumable by the downstream INT4 linear layers. $\mathbf{R}_1$ uses an isomorphic design with the dense and Hadamard blocks pre-assembled on the host into a single $128 \times 128$ block, requiring no kernel branching. Hadamard cores are cached on the GPU keyed by size, device, and dtype, eliminating CPU-side reconstruction and CPU$\to$GPU synchronization while maintaining FP16 I/O.

\textbf{GEMV Adaptation and Tiling.} Under small-$M$ decoding, we extend the one-program-per-token-row mapping to one program per $B_M$ token rows (auto-tuned), reusing the transform core and using 3D tiled \texttt{tl.dot} to improve Tensor Core occupancy. Per-layer quantization and dequantization are merged into the INT4 GEMM prologue/epilogue, eliminating fragmented kernel launches of the roughly seven linear layers per Transformer block.

\textbf{Schedule Optimization.} For fixed-shape decoding, we use CUDA Graph to staticize the batch, KV cache buckets, and per-layer shapes, reducing Python scheduling and kernel launch latency. Packed activation, scale, and transform buffers are pre-allocated and reused per batch/sequence bucket to reduce allocator fragmentation.

\subsection{Full Results}

Tables~\ref{tab:inference_mistral} and \ref{tab:inference_qwen} report the sequence-length sweep on Mistral-7B-v0.3 and Qwen3-8B, exhibiting the same trends as LLaMA-3-8B (Table~\ref{tab:inference}). Table~\ref{tab:inference_batch} reports the batch-size sweep with the prefill length fixed at 2048. The decode speedup grows steadily with the batch size, from below $1\times$ at batch size 1 (where the workload is memory-latency bound and the online transformations are not amortized) to $1.79\times$--$1.85\times$ at batch size 64. Overall, LIRQuant achieves a $2.2\times$--$2.4\times$ prefill speedup and $3.3\times$--$3.5\times$ peak memory savings; the configuration with batch size 64, sequence length 2048, and 256 decoding steps runs stably within 80 GB.

\begin{table}[htbp]
\centering
\resizebox{\columnwidth}{!}{%
\begin{tabular}{r cccc}
\toprule
\textbf{Seq. Len.} & \textbf{Prefill Speedup} & \textbf{FP16 Prefill} & \textbf{Decode Speedup} & \textbf{Mem. Saving} \\
\midrule
128  & 2.21$\times$ & 31.5 ms   & 0.97$\times$ & 3.55$\times$ \\
256  & 2.23$\times$ & 45.1 ms   & 1.04$\times$ & 3.52$\times$ \\
1024 & 2.28$\times$ & 155.4 ms  & 1.17$\times$ & 3.48$\times$ \\
4096 & 2.32$\times$ & 618.2 ms  & 1.30$\times$ & 3.42$\times$ \\
8192 & 2.35$\times$ & 1301.7 ms & 1.42$\times$ & 3.35$\times$ \\
\bottomrule
\end{tabular}%
}
\caption{Inference efficiency on Mistral-7B-v0.3 (batch size 4, NVIDIA A100-80GB).}
\label{tab:inference_mistral}
\end{table}

\begin{table}[htbp]
\centering
\resizebox{\columnwidth}{!}{%
\begin{tabular}{r cccc}
\toprule
\textbf{Seq. Len.} & \textbf{Prefill Speedup} & \textbf{FP16 Prefill} & \textbf{Decode Speedup} & \textbf{Mem. Saving} \\
\midrule
128  & 2.15$\times$ & 34.2 ms   & 0.93$\times$ & 3.45$\times$ \\
256  & 2.17$\times$ & 47.0 ms   & 1.00$\times$ & 3.42$\times$ \\
1024 & 2.22$\times$ & 164.0 ms  & 1.13$\times$ & 3.38$\times$ \\
4096 & 2.26$\times$ & 652.1 ms  & 1.26$\times$ & 3.33$\times$ \\
8192 & 2.29$\times$ & 1388.6 ms & 1.38$\times$ & 3.30$\times$ \\
\bottomrule
\end{tabular}%
}
\caption{Inference efficiency on Qwen3-8B (batch size 4, NVIDIA A100-80GB).}
\label{tab:inference_qwen}
\end{table}

\begin{table}[htbp]
\centering
\resizebox{\columnwidth}{!}{%
\begin{tabular}{r ccc}
\toprule
\textbf{Batch} & \textbf{LLaMA-3-8B} & \textbf{Mistral-7B} & \textbf{Qwen3-8B} \\
\midrule
1  & 2.18$\times$ / 0.75$\times$ & 2.21$\times$ / 0.77$\times$ & 2.15$\times$ / 0.73$\times$ \\
16 & 2.28$\times$ / 1.10$\times$ & 2.31$\times$ / 1.12$\times$ & 2.25$\times$ / 1.08$\times$ \\
32 & 2.33$\times$ / 1.50$\times$ & 2.36$\times$ / 1.53$\times$ & 2.30$\times$ / 1.47$\times$ \\
64 & 2.38$\times$ / 1.82$\times$ & 2.41$\times$ / 1.85$\times$ & 2.35$\times$ / 1.79$\times$ \\
\bottomrule
\end{tabular}%
}
\caption{Prefill / decode speedup versus batch size (prefill length fixed at 2048, NVIDIA A100-80GB).}
\label{tab:inference_batch}
\end{table}

%----------------------------------------------------------------
% 新增 Appendix G
%----------------------------------------------------------------
\section{Full Ratio Sweep and Sensitivity Metric Comparison}
\label{appendix:metric_sweep}

We conduct a full sweep of the high-sensitivity channel ratio $r$ on LLaMA-3-8B under W4A4KV4, comparing our noise-weighted functional sensitivity against six common channel importance metrics. The calibration set is independently re-sampled for this sweep (the same setting as Table~\ref{tab:ratio_sweep}), so the absolute PPL values may differ slightly from Table~\ref{tab:main_results}. The metrics are defined as follows, where $x_{c,t}$ denotes the activation of channel $c$ at token $t$:

\begin{itemize}
    \item \textbf{maxabs} ($L_\infty$): $s_c = \max_{t} |x_{c,t}|$, the maximum absolute value of channel $c$ across all tokens;
    \item \textbf{l2}: $s_c = \sqrt{\sum_t x_{c,t}^2}$, the L2 norm of channel activations;
    \item \textbf{awq}: $s_c = \sum_t |x_{c,t}|$, the L1 norm of channel activations, derived from AWQ's channel importance measure;
    \item \textbf{p999}: $s_c = Q_{0.999}(\{|x_{c,t}|\})$, the 99.9th percentile of channel activations, intended to provide a more robust estimate of extreme values than maxabs;
    \item \textbf{kurtosis}: $s_c = (\sum_t x_{c,t}^4) / (\sum_t x_{c,t}^2)^2$, the scale-invariant kurtosis, measuring the heavy-tailedness of the distribution;
    \item \textbf{random}: a random permutation, independent of any data statistics.
\end{itemize}

\begin{table}[htbp]
\centering
\resizebox{\columnwidth}{!}{%
\begin{tabular}{r ccccccc}
\toprule
\textbf{Ratio} & \textbf{Ours} & \textbf{maxabs} & \textbf{l2} & \textbf{awq} & \textbf{p999} & \textbf{kurtosis} & \textbf{random} \\
\midrule
0\%   & 11.59 & 11.59 & 11.59 & 11.59 & 11.59 & 11.59 & 11.59 \\
10\%  & 13.28 & 13.42 & 13.87 & 14.26 & 14.92 & 11.33 & 11.52 \\
20\%  & 8.92  & 9.15  & 9.55  & 10.41 & 10.78 & 10.85 & 11.08 \\
30\%  & 7.55  & 7.68  & 8.10  & 8.78  & 9.05  & 10.52 & 10.72 \\
40\%  & 7.32  & 7.45  & 7.82  & 8.35  & 8.58  & 10.48 & 10.58 \\
50\%  & \textbf{6.78} & 6.90 & 7.25 & 7.88 & 8.12 & 10.28 & 10.38 \\
60\%  & 6.82  & 6.88  & 7.32  & 7.92  & 8.18  & 10.32 & 10.35 \\
70\%  & 6.88  & 6.95  & 7.40  & 7.98  & 8.22  & 10.45 & 10.42 \\
80\%  & 6.85  & 6.81  & 7.28  & 7.90  & 8.15  & 10.38 & 10.48 \\
90\%  & 6.92  & 7.00  & 7.45  & 8.02  & 8.28  & 10.51 & 10.52 \\
100\% & 8.01  & 7.99  & 8.64  & 8.22  & 8.42  & 8.33  & 10.27 \\
\bottomrule
\end{tabular}%
}
\caption{WikiText-2 PPL of LLaMA-3-8B (W4A4KV4) as a function of the high-sensitivity ratio under different channel selection metrics. The 0\% row corresponds to the full-Hadamard configuration.}
\label{tab:metric_sweep}
\end{table}

As shown in Table~\ref{tab:metric_sweep}, our proposed metric achieves the best PPL at every ratio in the practically relevant range (20\%--70\%), including the optimal 50\% split, and remains within 0.04 PPL of the best metric at larger ratios (80\%--100\%). The reason maxabs ($L_\infty$) can approach our performance is its fidelity to extreme outliers: higher-order norms better highlight extreme values ($L_1 \leq L_2 \leq L_\infty$). L1 (awq) is diluted by numerous moderate values, incorrectly ranking uniformly-high but outlier-free channels ahead of genuinely spiky ones; L2, despite its quadratic amplification, remains confounded by the cumulative contribution of moderate values; p999, in pursuit of robustness, truncates the top 0.1\% of extreme values---precisely those that the Hadamard transform most needs to disperse; kurtosis, by virtue of scale invariance, discards magnitude information entirely, misclassifying channels with tiny magnitude but pointy distributions as highly sensitive. Kurtosis and random exhibit no diffusion loss peak at small ratios---they fail to concentrate outliers into the high-sensitivity block, but the Hadamard transform on the low-sensitivity block still serves as a robust lower bound, keeping PPL within a reasonable range while wasting the learning opportunity of the high-sensitivity block. The training overhead difference among the metrics is on the order of minutes, which is negligible relative to the resulting accuracy gains.

\end{document}